\pdfoutput=1

\documentclass[11pt]{article}

\usepackage{booktabs}

\usepackage[final]{acl}

\usepackage{times}
\usepackage{latexsym}

\usepackage[T1]{fontenc}

\usepackage[utf8]{inputenc}

\usepackage{microtype}

\usepackage{inconsolata}

\usepackage{graphicx}

\usepackage{adjustbox}
\usepackage{array}
\usepackage{amsmath}
\usepackage{cleveref}
\usepackage{subcaption}
\usepackage{placeins}

\title{Hakim: Farsi Text Embedding Model}

\author{%
  Mehran Sarmadi$^{\ddagger\mathsection}$,
  Morteza Alikhani$^{\ddagger\mathsection}$,
  Erfan Zinvandi$^{\mathsection}$,
  Zahra Pourbahman$^{\mathsection}$\\
  $^{\ddagger}$MCINEXT, $^{\mathsection}$Sharif University of Technology\\
  \texttt{\{mehran.sarmadi99, morteza.alikhani95, e.zeynvandi1376\}@sharif.edu, z.pourbahman@shahed.ac.ir}
}

\begin{document}
\maketitle
\begin{abstract}
Recent advancements in text embedding have significantly improved natural language understanding across many languages, yet Persian remains notably underrepresented in large-scale embedding research. In this paper, we present Hakim, a novel state-of-the-art Persian text embedding model that achieves a 8.5\% performance improvement over existing approaches on the FaMTEB benchmark, outperforming all previously developed Persian language models. As part of this work, we introduce three new datasets—Corpesia, Pairsia-sup, and Pairsia-unsup—to support supervised and unsupervised training scenarios. Additionally, Hakim is designed for applications in chatbots and retrieval-augmented generation (RAG) systems, particularly addressing retrieval tasks that require incorporating message history within these systems. We also propose a new baseline model built on the BERT architecture. Our language model consistently achieves higher accuracy across various Persian NLP tasks, while the RetroMAE-based model proves particularly effective for textual information retrieval applications. Together, these contributions establish a new foundation for advancing Persian language understanding.

\end{abstract}

\section{Introduction}

Text embedding models play a pivotal role in modern Natural Language Processing (NLP) by transforming textual data into numerical representations that capture semantic and contextual meaning. While languages such as English have benefited from extensive research in embedding methodologies, Persian remains significantly underrepresented. Existing multilingual models, such as Multilingual E5~\cite{wang2024multilinguale5textembeddings} and BGE-M3~\cite{chen-etal-2024-m3}, struggle to capture the intricacies of Persian grammar and semantics due to the limited availability of high-quality training data.

In training a high-quality text embedding model, the choice of a strong foundational language model plays a crucial role. Several pre-trained language models have been developed for Persian, including ParsBERT~\cite{Farahani2020ParsBERTTM}, FaBERT~\cite{Masumi2024FaBERTPB}, and TookaBERT~\cite{sadraeijavaheri2024tookabertstepforwardpersian}. These models utilize the BERT algorithm to train their language representations. However, traditional BERT-based models do not explicitly optimize for embedding generation. Since text embedding models aim to capture the semantic meaning of a given text within the CLS token, models like the RetroMAE~\cite{xiao-etal-2022-retromae}, which apply a dedicated loss function on the CLS token during training, are better suited for this task. Therefore, there is a clear need to train a new foundational language model specifically optimized for text embeddings.

One of the most critical aspects of training a foundational language model is ensuring access to a clean and diverse dataset. To construct such a dataset, we crawled a wide range of Persian-language websites across different domains. We then individually processed and denoised the text from each website to curate a high-quality corpus, which we named Corpesia. This dataset comprises text from 46 websites across 21 broad topics and contains over 11 billion tokens. We refer to this dataset as Corpesia, and utilize it alongside two other datasets, hmBlogs and Queries, for training two models: BERT and RetroMAE.

To train a high-quality text embedding model, having a large collection of semantically paired texts is essential for understanding the relationship between two pieces of text. To construct a robust and comprehensive dataset, we collected a large corpus of 50 million text pairs. Following a series of filtering steps, this collection was refined and reduced to 5 million high-quality pairs, resulting in the final dataset, referred to as Pairsia-unsup. This dataset encompasses a wide range of topics and tasks within the Persian language, ensuring diverse and representative coverage. By incorporating various linguistic structures, contextual meanings, and semantic relationships, our dataset aims to enhance the performance and generalizability of the text embedding model. The inclusion of such a vast and varied corpus is essential for capturing the intricacies of Persian text and improving the model’s ability to generate meaningful and contextually rich embeddings.

In the final phase of model training, we employed a carefully curated dataset comprising 1.3 million supervised instances. The construction of this dataset, referred to as Pairsia-sup, was guided by the objective of incorporating a diverse range of NLP tasks that are well-suited to text embedding-based solutions. This dataset served as the foundation for supervised training of the model. Moreover, task-specific instructions were integrated during this stage to effectively guide the model across different NLP tasks, leading to a substantial increase in performance, with accuracy improving by up to 5.71\%.

FaMTEB~\cite{zinvandi2025famtebmassivetextembedding} is a benchmark developed based on MTEB for evaluating text embedding models in the Persian language. This benchmark comprises 7 tasks and 63 datasets aimed at assessing the performance of text embedding models. One of the novel aspects highlighted in this benchmark is the evaluation of these models in tasks related to chatbots and retrieval-augmented generation (RAG).

In this work, we utilize the FaMTEB benchmark to evaluate our model, \textit{Hakim}. Additionally, for training the model on chatbot- and RAG-related tasks, we leverage the training data provided within this benchmark in the supervised training stage. The use of this data enables our model to support functionalities such as search with follow-up capability.

We introduce a novel task on which our model is trained, referred to as cross classification. This task is inspired by the architecture of Cross-Encoders. In cross classification, we concatenate a pair of texts with an instruction that reflects the semantic relationship between them, drawn from tasks such as retrieval, classification, semantic textual similarity (STS), or named entity recognition (NER). This augmented pair is then provided to the model as input. The label, in turn, specifies the nature of the relationship—e.g., whether the document answers the query (retrieval), whether the text belongs to a given class (classification), the similarity score between two sentences (STS), or the entity type of a highlighted token in the sentence (NER).

After training, our model achieves an average accuracy of 88\% on this task. One of the potential applications of this capability is the verification of LLM outputs across a range of tasks, enabling more reliable and interpretable responses.

Our main contributions are as follows:
\begin{itemize}
    \item Introduction of three new datasets—\textit{Corpesia}, \textit{Pairsia-sup}, and \textit{Pairsia-unsup}—used in the pretraining phase and for training the text embedding model.
    \item Development of a state-of-the-art text embedding model for the Persian language.
    \item Enabling the model's integration into RAG systems and chatbot applications.
    \item Demonstrating the model's applicability to the novel task of cross-classification.
\end{itemize}


\section{Related Work}
\subsection{General Text Embedding Approaches} 
Text embeddings have evolved from static representations such as Word2Vec~\cite{mikolov2013efficientestimationwordrepresentations} and FastText~\cite{bojanowski-etal-2017-enriching} to deep learning-based contextualized models like ELMo~\cite{peters-etal-2018-deep} and transformer-based architectures such as BERT and its variants. These advances have greatly improved the ability to capture rich semantic meanings across different linguistic contexts.

Recent breakthroughs in contrastive learning, particularly in models like SimCSE~\cite{gao-etal-2021-simcse} and Sentence-BERT~\cite{reimers-gurevych-2019-sentence}, have significantly advanced general-purpose text representations. However, models like SimCSE were primarily trained on single tasks such as sentence similarity and are not inherently suitable for broader applications like classification, question answering, and other complex downstream tasks.


A new generation of text embedding models has been developed to address the need for models that generalize across a wider range of tasks. These models are designed to produce representations that are more adaptable and robust for diverse NLP applications, such as retrieval, classification, and QA, thereby overcoming the limitations of earlier single-task-trained embeddings.

Among these models, the Bilingual Generative Embeddings model(BGE)~\cite{10.1145/3626772.3657878} has demonstrated significant advancements in text representation through a two-step contrastive training process. In the first step, BGE was trained on a large-scale dataset comprising 100 million unlabeled paired samples, which were widely collected from the web and structured data sources across various domains. This extensive pretraining phase allowed the model to develop a broad understanding of linguistic structures and semantic relationships.

In the second step, BGE was fine-tuned using a dataset containing 838,465 labeled samples, covering a diverse range of NLP tasks to enhance its ability to generate high-quality text embeddings. These tasks were specifically designed to assess and improve different aspects of text representation, ensuring the model's robustness and adaptability. As a result of this two-stage training approach, BGE achieved state-of-the-art performance in English and Chinese language processing, demonstrating superior effectiveness in multiple NLP benchmarks.

Many other text embedding models, such as GTE and Nomic~\cite{nussbaum2025nomicembedtrainingreproducible}, apply a similar strategy to train general-purpose text embedding models. These approaches leverage large-scale contrastive training and fine-tuning on diverse labeled datasets to enhance their performance across various languages and tasks.

In addition to these approaches, NV-Embed~\cite{lee2025nvembedimprovedtechniquestraining} introduces a high-performance text embedding model based on decoder-only architectures. Unlike traditional methods that rely on <EOS> token representations or mean pooling, NV-Embed employs a latent attention layer to generate more effective text embeddings.

The model follows a two-step training process: first, it is pretrained on a large-scale retrieval dataset to learn query-document relationships; second, it is fine-tuned on diverse tasks such as classification and clustering to enhance generalization. Additionally, NV-Embed incorporates instruction-based learning to optimize performance across specific NLP tasks, demonstrating the potential of decoder-only models for robust text representation.

\subsection{Persian Text Embedding}
In the Persian language, the only publicly available text embedding model to date is Tooka SBERT, for which detailed training information is not available. Among multilingual models that support Persian, notable examples include BGE-m3, Jina-embed-v3~\cite{sturua2024jinaembeddingsv3multilingualembeddingstask}, and mE5.

Although these models can generate vector representations for Persian text, Persian is not one of the primary languages used in their training. For example, Jina-embed-v3 supports 100 languages, but its primary focus is on 30 languages, excluding Persian. As a result, these models may not necessarily achieve optimal performance in Persian text representation.

\subsection{FaMTEB}
The Massive Text Embedding Benchmark (MTEB) is a comprehensive suite designed to evaluate text-embedding models across multiple NLP tasks, including classification, clustering, pair classification, reranking, retrieval, summarization, bitext mining, and semantic textual similarity (STS). MTEB has become the most popular benchmark for evaluating text embeddings. While MTEB primarily focuses on English, it lacks sufficient support for low-resource languages like Persian. To address this, FaMTEB extends MTEB by introducing a large-scale Persian benchmark with 63 datasets covering six of the same tasks as MTEB. However, instead of bitext mining and summarization, FaMTEB evaluates models on summary retrieval.

\begin{figure*}[t]
\centering
\includegraphics[width=\textwidth]{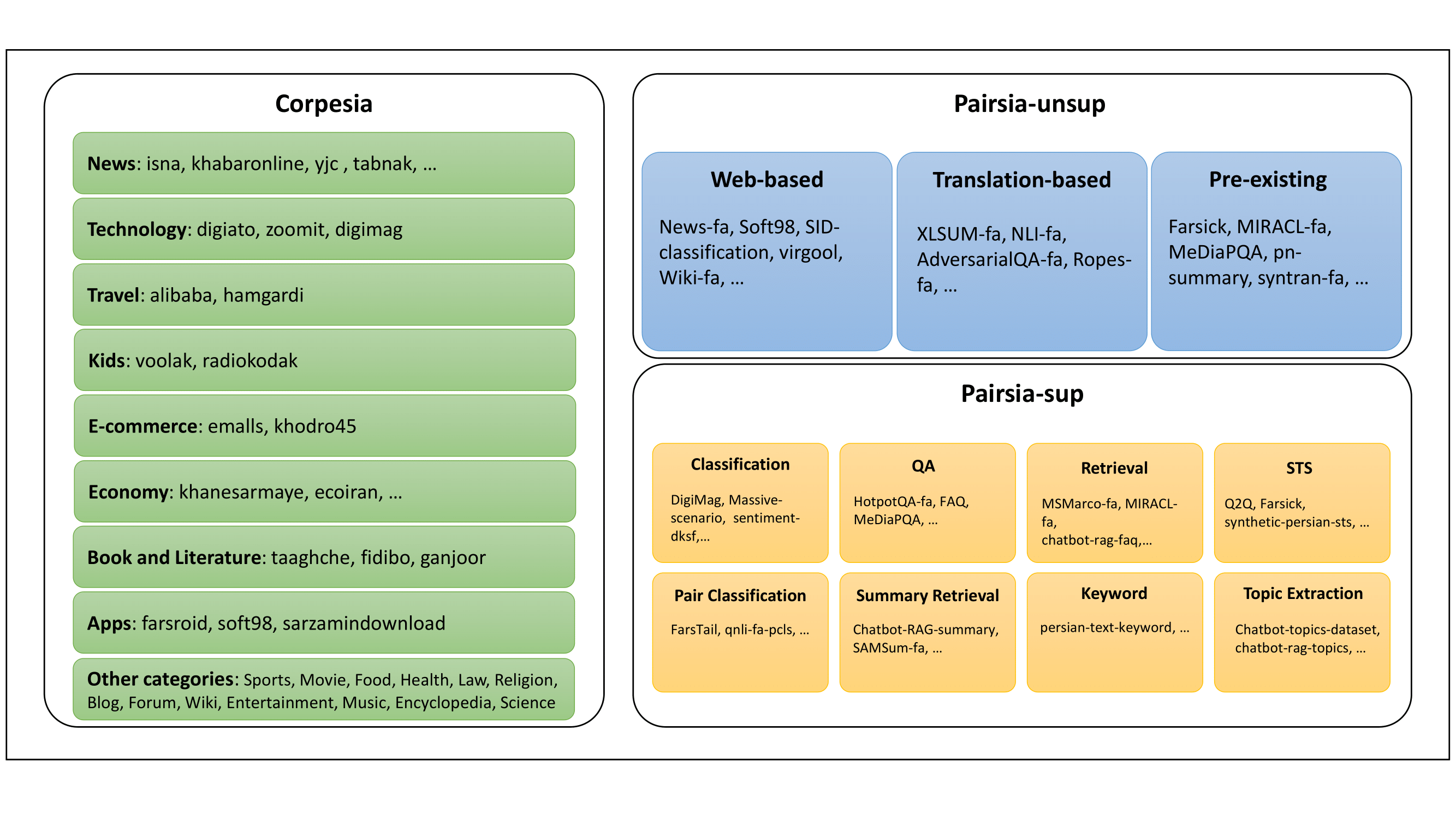}
\caption{This figure presents an overview of all the datasets utilized for training the Hakim network. The Corpesia corpus is employed for training the base model. The Pairsia-unsup dataset is used during the unsupervised training phase, while the Pairsia-sup dataset is incorporated in the final stage of training.}
\label{fig:hakim}
\end{figure*}

\section{Method}
\subsection{Data}
\subsubsection{Pre-training Data}
\textbf{Corpesia}  This dataset comprises textual data from Iranian websites, categorized into 21 distinct categories—such as news, economy, and books, as illustrated in Table~\ref{tab:train_count}—and sourced from 46 websites. Each website, after being crawled using a dedicated crawler, is extracted as a structured data instance, preserving the hierarchical layout of titles, headings, and paragraphs in the stored file. We used \textit{Selectolax}, a fast HTML parser, to efficiently parse the HTML content during this process. In addition, various forms of textual noise—such as advertisements, source citations, and site-specific template phrases—are identified and filtered separately.

To prepare data for pretraining \textit{HakimBERT}, we used a probabilistic strategy to combine text, rather than treating each paragraph as a separate sample. Specifically, 90\% of webpages were dynamically merged into longer text blocks until the model token limit was reached. The remaining 10\% followed the standard approach, with paragraphs kept separate. In the merged set, document headings were retained in 50\% of cases to preserve structural cues. Additionally, 10\% of merged samples were allowed to slightly exceed the model token limit before being split, helping the model generalize better to longer sequences while still respecting practical training limits.

Overall, the dataset amounts to approximately 11 billion tokens, which is significantly smaller than larger datasets like Targoman~\cite{targoman_pws} (around 41 billion tokens). However, as discussed in Section \ref{sec:analysis}, pretraining our base model on this dataset led to better downstream results compared to TookaBERT, which is trained on Targoman and two other datasets.

\begin{table}[ht]
    \begin{adjustbox}{width=\columnwidth}
    \begin{tabular}{c|c|c}
        Category & Web Urls Domains & \#tokens \\
        \hline
        & isna.ir, khabaronline.ir, tasnimnews.com,\\
        News & yjc.ir, hamshahrionline.ir, mehrnews.com & 4,748,358,290 \\
        & tabnak.ir, asriran.com \\
        \hline
        Technology & digiato.com, zoomit.ir, digikala.com/mag, & 414,639,633 \\
        \hline
        Science & bigbangpage.com & 8,786,319 \\
        \hline
        Economy & khanesarmaye.com, ecoiran.com, donya-e-eqtesad.com & 822,676,697 \\
        \hline
        Books \& Literature & taaghche.com, fidibo.com, ganjoor.net & 162,292,381 \\
        \hline
        Apps & farsroid.com, soft98.ir, sarzamindownload.com & 17,185,689 \\
        \hline
        Health & newmiind.com, doctoreto.com & 35,301,238 \\
        \hline
        Sports & varzesh3.com & 99,979,068 \\
        \hline
        Music & upmusics.com, music-fa.com & 15,745,744 \\
        \hline
        Movie & vipofilm.com & 4,533,569 \\
        \hline
        E-commerce & emalls.ir, khodro45.com & 46,405,081 \\
        \hline
        Travel & hamgardi.com, alibaba.ir/mag & 34,242,853 \\
        \hline
        Food & parsiday.com & 6,658,426 \\
        \hline
        Kids & voolak.com, radiokodak.com & 2,084,614 \\
        \hline
        Religion & wiki.ahlolbait.com, hawzah.net, fa.wikishia.net & 461,472,962 \\
        \hline
        Blog & virgool.io, magerta.ir, motamem.org & 669,092,816 \\
        \hline
        Forum & ninisite.com & 977,392,614 \\
        \hline
        Wiki & fa.wikipedia.org & 2,132,566,416 \\
        \hline
        Encyclopedia & abadis.ir & 127,148,347 \\
        \hline
        Entertainment & beytoote.com, namnak.com & 375,829,131 \\
        \hline
        Law & wikihoghoogh.net & 3,369,729 \\
        \hline
        SUM & - & 11,165,761,617 \\
    \end{tabular}
    \end{adjustbox}
    \caption{The available data in Corpesia has been presented in detail in this table.}
    \label{tab:train_count}
\end{table}

\textbf{hmBlogs} The hmBlogs dataset is a large-scale Persian corpus built from approximately 20 million blog posts, containing around 6.8 billion tokens.

\textbf{Queries} This dataset consists of 8.5 million anonymized queries collected from a Persian-language search engine.

\subsubsection{Pairsia-unsup}
The dataset comprises diverse types of pairs, including title-document, question-answer, FAQ-style pairs, and more. We construct this dataset through three primary sources: (1) structured text extracted from websites, (2) machine-translated English corpora, and (3) Persian datasets that already existed.

Due to the inherent noise in the collected data, we apply the BGE-M3 text embedding model for data cleaning. Given the varying distributions across datasets and the fact that the embedding model yields different similarity scores across domains, we compute a separate similarity threshold for each subset and filter accordingly. We also balance the datasets across different task types to prevent the model from becoming biased toward any particular task. This process reduces the unsupervised dataset from 50 million pairs to 5 million processed pairs.

In the following, several of the datasets used in this study are described.

\textbf{News-fa}  
This dataset consists of pairs collected from news websites. The pairs are either formed by matching the page title with the page summary (as most news sections include a summary for each page), or by pairing the page summary with the first paragraph of the page. The news websites used to construct this dataset include ten sources: Asriran, Donya-e-Eqtesad, Ecoiran, Hamshahri Online,ISNA, Khabar Online, Tabnak, Tasnim, Varzesh3, and YJC.

\textbf{Web Page}  Most websites follow a general structure as shown below:
\[
    \{\text{H}_1, \text{P}, \text{H}_2, \text{P}, \text{H}_2, \text{P}, \ldots\}
\]
Here, \(\text{H}_1\) represents the main title of the page, followed by a general explanation or overview P. This is typically followed by several subsections, each introduced by \(\text{H}_2\), which denotes a specific topic, accompanied by corresponding passages.  

We construct pairs from various websites across different domains by forming pairs such as \((\text{H}_1, \text{P})\) and \((\text{H}_2, \text{P})\). Examples of such websites include khanesarmaye, javab24, farsroid, and parsiday, which cover a wide range of topics, including economics, cooking, technology, etc.

\textbf{FAQ}  This dataset comprises frequently asked questions (FAQs) and their corresponding answers collected from various organizations and companies, including mobile network operators, energy providers, and others.

\textbf{Papers}  We also constructed a collection of title–abstract pairs from papers by crawling two academic websites: SID and Irandoc.

\textbf{NLI-fa}  This dataset comprises machine-translated versions of the \textit{MNLI} and \textit{SNLI} datasets in Persian. Models trained on this dataset, such as SimCSE~\cite{gao-etal-2021-simcse}, are capable of producing high-quality embeddings.

\textbf{Other translated datasets}  To train the model, several valuable English datasets—whose equivalents are scarce in Persian—were machine-translated into Persian. These datasets include \textit{AdversarialQA}, which consists of questions, context passages, and short answers derived from the context;  \textit{MS MARCO}, consisting of query–passage pairs based on real user queries from the Bing search engine; \textit{QNLI}, a collection of question-answer pairs each labeled to indicate whether the answer correctly addresses the question; \textit{ROPES}, which evaluates whether a conclusion can be inferred from a paragraph; \textit{SAMSum}, comprising dialogue-summary pairs; and \textit{WikiLingua}, collected from WikiHow, containing instructional texts paired with their summaries.

\textbf{Other existing datasets} In addition to our proprietary data, we also leveraged several publicly available datasets that have previously been used in various Persian text embedding tasks. Among these datasets are \textit{FarSICK}~\cite{9721521}, a Persian translation of the English SICK dataset for the Semantic Textual Similarity (STS) task; \textit{FarsTAIL}~\cite{DBLP:journals/corr/abs-2009-08820}, which contains data for the textual entailment task in Persian; The \textit{Persian Web Document Retrieval Corpus}~\cite{10553090} is a collection consisting of query-document pairs retrieved from the web; and \textit{SynTran-fa}~\cite{farsi2024syntran}, a collection of short question-answer pairs in Persian. We also utilized other datasets that are applicable to related text embedding tasks.

\subsubsection{Pairsia-sup}
For supervised fine-tuning, we curate a high-quality and diverse set of clean Persian data across various tasks relevant to text embedding, including retrieval, classification, semantic textual similarity (STS), question answering (QA), and others. To the best of our knowledge, this is the first Persian dataset of such scale and diversity in this domain.

Each example in Pairsia-sup is paired with nine negative samples. These include: (1) three hard negatives constructed from positive pairs within the same dataset, (2) three hard negatives sampled across all datasets, and (3) three random negatives.

To extract hard negatives, all relevant documents are indexed using the model trained on Pairsia-unsup. Then, for a given query, the top K retrieved documents are discarded, since these negative pairs may be too similar to the correct answer, and three negatives are randomly selected from the range between ranks K and L. The value of L serves as an upper bound for the selection of candidates. The parameters K and L are chosen separately for each dataset to control the difficulty level of the negative samples.

Some of the datasets included in Pairsia-Sup are:

\textbf{FaMTEB}
We utilize the major datasets provided in the training part of the FaMTEB datasets to address various tasks. These include the train split of \textit{BEIR-fa}, synthetic datasets, datasets specifically developed for chatbot applications, and the \textit{Query-to-Query} dataset, which brings semantically similar queries closer together.

\textbf{Other existing datasets} 
Some of the datasets used for supervised training overlap with those used in the unsupervised setting. However, the key differences in this stage are the cleanliness of the utilized data, the presence of sufficient examples for each task, and the fact that each pair also received multiple negatives, as explained before.

\subsection{Model}
\subsubsection{Pretraining}
To pretrain our base model and train a new tokenizer, we leveraged two rich datasets: our proprietary collection, \textbf{Corpesia}, and the publicly available \textbf{HmBlogs}~\cite{DBLP:journals/corr/abs-2111-02362}. In addition to these, we incorporated user queries sourced from an Iranian search engine, enhancing the model’s robustness to conversational language, misspellings, and incomplete intent expressions.
To train our tokenizer, we first preprocessed all datasets. The preprocessing pipeline consisted of the following steps:

\begin{itemize}
    \item \textbf{Character Filtering:} We removed characters not commonly used in Persian. Since the use of Arabic and Latin characters is prevalent in modern Persian text, we retained both. Additionally, we preserved emojis and punctuation marks, as they are widely used across languages and can carry semantic information.
    \item \textbf{Noise Removal:} We eliminated URLs and residual HTML tags from the text to reduce noise and improve the quality of the training data.
\end{itemize}

After preprocessing, we trained a tokenizer using the WordPiece algorithm with a vocabulary size of 50,000 tokens. For normalization, we applied the NFKC standard to simplify and unify the text. Furthermore, we normalized Arabic letters that have different forms in Persian to their standard Persian equivalents. Numerical digits were also converted to English numerals to ensure consistency throughout the corpus.

The data was fed to the model line-by-line for inputs exceeding 100 tokens. For shorter texts, we used a grouped-token strategy: each piece of text was tokenized and concatenated with others until reaching the model’s maximum token capacity.

To improve the capability of the model in dense retrieval and representation learning, we adopted a \textbf{Masked Autoencoder (MAE)} approach inspired by RetroMAE~\cite{xiao-etal-2022-retromae}. This method employs a lightweight decoder following the BERT encoder, specifically applied to the [CLS] token, to produce a more semantically enriched representation. 

\subsubsection{Stage 1 (Unsupervised)}
In this stage, we train the model on a large-scale corpus of semantically related text pairs, Pairsia-Sup, to enable it to acquire general-purpose semantic representations.

We pretrain the model using this corpus with a contrastive learning objective based on the InfoNCE loss. In this part of training, we bring the query closer to its positive pair, while using in-batch negatives for the negative samples. Further experimental details are provided in Section~\ref{sec:experiments}.

\subsubsection{Stage 2 (Supervised)}

In this stage of training, we fine-tune the model previously trained in Stage 1 using the Pairsia-sup dataset along with the corresponding instructions. Based on the specific task associated with each data instance, we append a distinct instruction to both elements of each pair. In general, we incorporate instructions into the dataset in three distinct forms: first, for classification tasks; second, for cross-classification tasks; and third, for cases where the two elements of the pair share a semantic relationship or form a relative pair (Figure \ref{fig:inst_structure}).

In classification tasks, the first element of the pair typically contains a piece of text that must be classified into a category represented by the second element. As illustrated in Figure \ref{fig:inst_structure}, we append an instruction to each element of the pair accordingly. In cross-classification tasks, the goal is to assess the relationship between two textual components embedded within a single instruction-driven prompt. For instance, in a question answering (QA) scenario, we assess whether a given question and answer—each framed within an instructional template—form a valid pair. In what we refer to as relative pair tasks, the second element of the pair bears a specific relationship to the first. This relationship may take the form of a question-answer pair, semantic relevance, a retrieved document, or similar associations.

To ensure flexibility, the model is also exposed to the same data without instructions, enabling it to generalize to both instruction-based and instruction-free usage scenarios.

Finally, the model is trained using a contrastive learning objective similar to the pretraining stage. Further experimental results and analysis are provided in Section~\ref{sec:experiments}.

\begin{table}[]
    \centering
    \begin{adjustbox}{width=\columnwidth}
    \begin{tabular}{>{\centering\arraybackslash}p{2cm}|>{\centering\arraybackslash}p{3cm}|>{\centering\arraybackslash}p{3cm}}
        & Stage 1 & Stage 2 \\
        \hline
        Datasets & digikala\_mag, farsroid, farsick, irandoc-pair, MSMARCO\_fa, news\_fa and etc & alpaca\_fa, HotpotQA\_fa, MeDiaPQA, MIRACL\_fa, SyntheticPersianQA, Q2Q and etc \\
        \hline
        Size & 5,371,643 & 1,302,659 \\
    \end{tabular}
    \end{adjustbox}
    \caption{Sample of datasets used in stage1 and stage2 of training.}
    \label{tab:train_count}
\end{table}

\subsubsection{Training Details}
\label{sec:training_details}
\textbf{Pretraining loss:} BERT is pretrained using the Masked Language Modeling (MLM) objective. Given a sequence of tokens $X = [x_1, x_2, \dots, x_n]$, a subset of tokens $\{x_{i_1}, x_{i_2}, \dots, x_{i_k}\}$ is randomly selected and replaced with a special \texttt{[MASK]} token. The goal is to predict the original tokens based on the masked input.

Let $\hat{x}_{i_j}$ be the model's prediction for the masked token $x_{i_j}$. The MLM loss is computed using cross-entropy over the vocabulary for each masked position:

\[
\mathcal{L}_{\text{MLM}} = - \sum_{j=1}^{k} \log P(\hat{x}_{i_j} = x_{i_j} \mid X_{\text{masked}})
\]

Here, $X_{\text{masked}}$ denotes the input sequence with selected tokens replaced by \texttt{[MASK]}. The model is trained to maximize the likelihood of the true tokens given their masked context, enabling it to capture deep bidirectional contextual representations.

The total training loss is averaged over all masked positions and batch elements. This self-supervised strategy allows BERT to learn rich language representations from large unlabeled corpora.

\textbf{RetroMAE loss:} 
RetroMAE enhances BERT-style pretraining by introducing a lightweight decoder and modifying the loss function to improve the representation of the \texttt{[CLS]} token for retrieval tasks. The model uses two distinct masking strategies:

1. \textbf{Encoding Stage}: A moderately masked version of the input, \( \tilde{X}_{\text{enc}} \), is processed by the encoder \( \Phi_{\text{enc}} \) to produce the sentence embedding \( h_{\tilde{X}} \):

   \[
   h_{\tilde{X}} \leftarrow \Phi_{\text{enc}}(\tilde{X}_{\text{enc}})
   \]

2. \textbf{Decoding Stage}: A more aggressively masked version, \( \tilde{X}_{\text{dec}} \), is combined with \( h_{\tilde{X}} \) to form the input for the decoder \( \Phi_{\text{dec}} \), which reconstructs the original sentence:

    \[
   H_{\tilde{X}_{\text{dec}}} = [h_{\tilde{X}}, e_{x_1} + p_1, \dots, e_{x_n} + p_n]
    \]

    \[
   \mathcal{L}_{\text{dec}} = \sum_{x_i \in \text{masked}} \text{CE}\left(x_i \mid \Phi_{\text{dec}}(H_{\tilde{X}_{\text{dec}}})\right)
   \]

 This setup forces the encoder to generate high-quality, semantically rich sentence embeddings, as the decoder alone cannot recover the full input without meaningful representations. This strategy not only increases the difficulty of the reconstruction task but also ensures that nearly all input tokens contribute learning signals, significantly boosting data efficiency.

\textbf{RetroMAE-v2 (DupMAE):} Introduces a duplex masked auto-encoder framework with two complementary decoding tasks:
\begin{itemize}
    \item Reconstructing the original input sentence based on the [CLS] embedding.
    \item Predicting the bag-of-words (BoW) features of the input sentence using the embeddings of the ordinary tokens. These two tasks are combined to train a unified encoder, enhancing the semantic representation capability by leveraging both [CLS] and token-level embeddings.

\end{itemize}

\textbf{InfoNCE loss:}
InfoNCE loss encourages similar (positive) text pairs to have closer representations while pushing dissimilar (negative) pairs apart. For each query $q$, a relevant document $d^+$, and a set of irrelevant documents $D^- = \{d_1^-, \ldots, d_n^-\}$, the InfoNCE loss is commonly used. It maximizes the similarity between $q$ and $d^+$ while minimizing it with respect to $D^-$. The loss is given by:

\[
\mathcal{L}_{cl} = -\log \frac{e^{s(q, d^+)/\tau}}{e^{s(q, d^+)/\tau} + \sum_{i=1}^{n} e^{s(q, d_i^-)/\tau}}
\]

where $s(\cdot, \cdot)$ is a similarity function like cosine and $\tau$ is a temperature parameter.

\section{Experiments}
\label{sec:experiments}

\subsection{Analysis}
\label{sec:analysis}
\textbf{BERT Like Pretraining:}
We pretrained our BERT-based model using the AdamW optimizer with hyperparameters $\beta_1 = 0.9$, $\beta_2 = 0.999$, and $\epsilon = 1\text{e}{-8}$. The learning rate was set to $5 \times 10^{-5}$, and we employed a linear learning rate scheduler with 1{,}000 warm-up steps. Pretraining was conducted for 3 epochs with a batch size of 32 per device. No weight decay was applied. Mixed-precision training (fp16) was enabled to improve computational efficiency. To ensure reproducibility, all experiments were run with a fixed random seed of 42.

\textbf{RetroMAE:} In the next stage, we continue pretraining the initial model on the same plain-text corpus using the RetroMAE-v2 loss for three epochs with a batch size of 64. Since text embedding models represent sentences solely through the [CLS] token, leveraging this model can substantially enhance the accuracy of downstream tasks.

\textbf{Unsupervised:} The initial dataset prepared for this stage consisted of 51,817,807 data pairs. After filtering out irrelevant entries and balancing samples, the dataset was reduced to 5,371,643 instances. A significant portion of the semantic pair data was collected from the web, primarily in the form of document-title pairs, which introduces a bias in the model toward this specific structure. To mitigate this bias, we perform sampling to reduce the overrepresentation of such instances.  We also experimented on various subsets of the dataset and chose the best dataset combination. Our model was trained for five epochs with a batch size of 16 on the filtered and rebalanced dataset.

Table~\ref{tab:faMTEB_results} presents a comparative evaluation of the Hakim-unsup model against other baseline models on the FaMTEB benchmark. As shown, a single stage of fine-tuning results in a substantial performance gain for the Hakim model, which is among the top 3 best-performing models by far.

\textbf{Supervised:} In the final stage of training, we utilize a total of 1.3M data pairs (which increases to approximately 4.5 million pairs after adding instructions in various forms) spanning various tasks included in the FaMTEB benchmark. The model is fine-tuned for 2 epochs with a batch size of 8, under two settings: with instruction tuning and without. Additionally, the instruction-tuned model is exposed to both instructed and non-instructed samples, enabling it to generalize across different usage scenarios.

The instruction-free model outperforms the previous best-performing model by 2.8\%, while the instruction-tuned variant achieves an even higher improvement of 8.5\%, as shown in Tables~\ref{tab:faMTEB_results} and \ref{tab:instruction_results}. These results highlight the substantial benefit of incorporating task instructions during fine-tuning, demonstrating that instruction tuning can significantly enhance model accuracy across a diverse range of tasks.

\begin{table*}[ht]
\centering
\label{tab:comparison1}
\resizebox{\textwidth}{!}{%
\begin{tabular}{l|cc|cc|cc|cc}
\toprule
& \multicolumn{2}{c|}{\textbf{SA}} & \multicolumn{2}{c|}{\textbf{QA}} & \multicolumn{2}{c|}{\textbf{NER}} & \multicolumn{2}{c}{\textbf{NLI}}  \\
\cmidrule(r){2-3} \cmidrule(r){4-5} \cmidrule(r){6-7} \cmidrule(r){8-9}
\textbf{Model} &  \textbf{DeepSentiPers} & \textbf{MirasOpinion} & \textbf{PQuAD} & \textbf{PCoQA} & \textbf{ParsTwiner} & \textbf{MULTICONER V2} & \textbf{FarsTail} & \textbf{ParsiNLU QP} \\
\midrule
BERT           & 62.11 & 81.03 & 71.47 & 19.24 & 57.67 & 30.53 & 67.24 & 72.62 \\
mBERT          & 67.21 & 83.71 & 86.25 & 47.25 & 76.95 & 56.88 & 82.14 & 79.43 \\
XLM-RoBERTa    & 76.35 & 85.05 & 87.87 & \textbf{47.36} & 80.45 & 53.47 & 84.53 & 78.72 \\
ParsBERT       & 78.03 & 84.68 & 86.67 & 42.18 & 82.69 & 60.31 & 82.38 & 79.45 \\
FaBERT         & 78.77 & \textbf{85.33} & 87.22 & 42.98 & 84.55 & 51.08 & 83.64 & 81.75 \\
TookaBERT-Base & 78.09 & 84.83 & 87.88 & 45.57 & 84.54 & 61.20 & 83.04 & 79.79 \\
\midrule
HakimBERT & \textbf{82.99} & 85.04 & \textbf{88.31} & 45.03 & \textbf{86.11} & \textbf{61.42} & \textbf{86.82} & \textbf{83.53} \\
\bottomrule
\end{tabular}
}
\caption{F1-score comparison of BERT-based models on eight Persian natural language understanding tasks spanning Sentiment Analysis (SA), Question Answering (QA), Named Entity Recognition (NER), and Natural Language Inference (NLI).}
\label{tab:base_model_eval}
\end{table*}

\begin{table*}[h!]
\centering
\begin{adjustbox}{width=\textwidth}
\begin{tabular}{c|cc|ccccccc}
\textbf{} & Size\,(M) & Avg. & Class. & Cluster. & PairClass. & Rerank. & Retriv. & STS & SumRet. \\
\hline

Tooka-SBERT                                   & 353 & 60.65 & 59.40 & 56.45 & 87.04 & 58.29 & 27.86 & 76.42 & 59.06 \\
GTE-multilingual-base                         & 305 & 63.64 & 56.07 & 57.28 & 84.58 & 69.72 & 41.22 & 75.75 & 60.88 \\
multilingual-e5-base                          & 278 & 62.93 & 57.62 & 56.52 & 84.04 & 72.07 & 41.20 & 74.45 & 54.58 \\
multilingual-e5-large                         & 560 & 64.40 & 59.86 & 57.19 & 84.42 & 74.34 & 42.98 & 75.38 & 56.61 \\
BGE-m3                                        & 567 & 65.29 & 58.75 & 57.73 & 85.21 & \textbf{74.56} & 43.38 & 76.35 & 61.07 \\
Jina-embeddings-v3                            & 572 & 64.53 & 59.93 & 59.15 & 83.71 & 61.26 & \textbf{43.51} & \textbf{78.65} & 65.50 \\ 
\midrule
Hakim-unsup.                               & 124 & 64.56 & 60.65 & 58.89 & 86.41 & 67.56 & 37.71 & 79.36 & 61.34  \\ 
Hakim-small.                               & 38 & 70.45 & 80.19 & 66.31 & 87.41 & 67.30 & 38.05 & 75.53 &  78.40 \\ 
Hakim                                      & 124 & \textbf{73.81} & \textbf{84.56} & \textbf{70.46} & \textbf{89.75} & 69.46 & 40.43 & 76.62 & \textbf{85.41} \\ 
\end{tabular}
\end{adjustbox}
\caption{Evaluation of various text embedding models on the FaMTEB benchmark.}
\label{tab:faMTEB_results}
\end{table*}

\subsection{Ablation Study}
\subsubsection{Zero Shot}
Since the model has been trained on diverse tasks with various instructions, it possesses the ability to perform zero-shot on different problems. To assess the zero-shot capabilities of our model, we evaluated it on a set of benchmark datasets that were not seen during training. This evaluation aims to examine the model's ability to generalize to new tasks and domains without any additional fine-tuning.

As shown in Table~\ref{tab:zeroshot_comparision}, the model achieves strong classification performance on the MassiveIntentClassification and HamshahriClustring datasets, despite having no prior exposure to similar data during training.

These results highlight the model’s effective generalization and support its ability to perform well in zero-shot scenarios.

\begin{table}[ht]
\centering
\begin{adjustbox}{width=\columnwidth}
\begin{tabular}{c|cc}

\textbf{Model Name} & MassiveIntentClassification & HamshahriClustring \\
\hline    
    tooka-sbert & 63.19 &  63.28 \\
    intfloat-multilingual-e5-large & 65.49 &  67.42 \\
    Hakim & 72.72 & 69.17  \\

\end{tabular}
\end{adjustbox}
\caption{Zero-shot performance on unseen benchmarks, showing strong generalization across tasks.}
\label{tab:zeroshot_comparision}
\end{table}

\subsubsection{Different Instructions}
\label{sec:different_inst}
\begin{figure}[ht]
\centering
\includegraphics[width=\columnwidth]{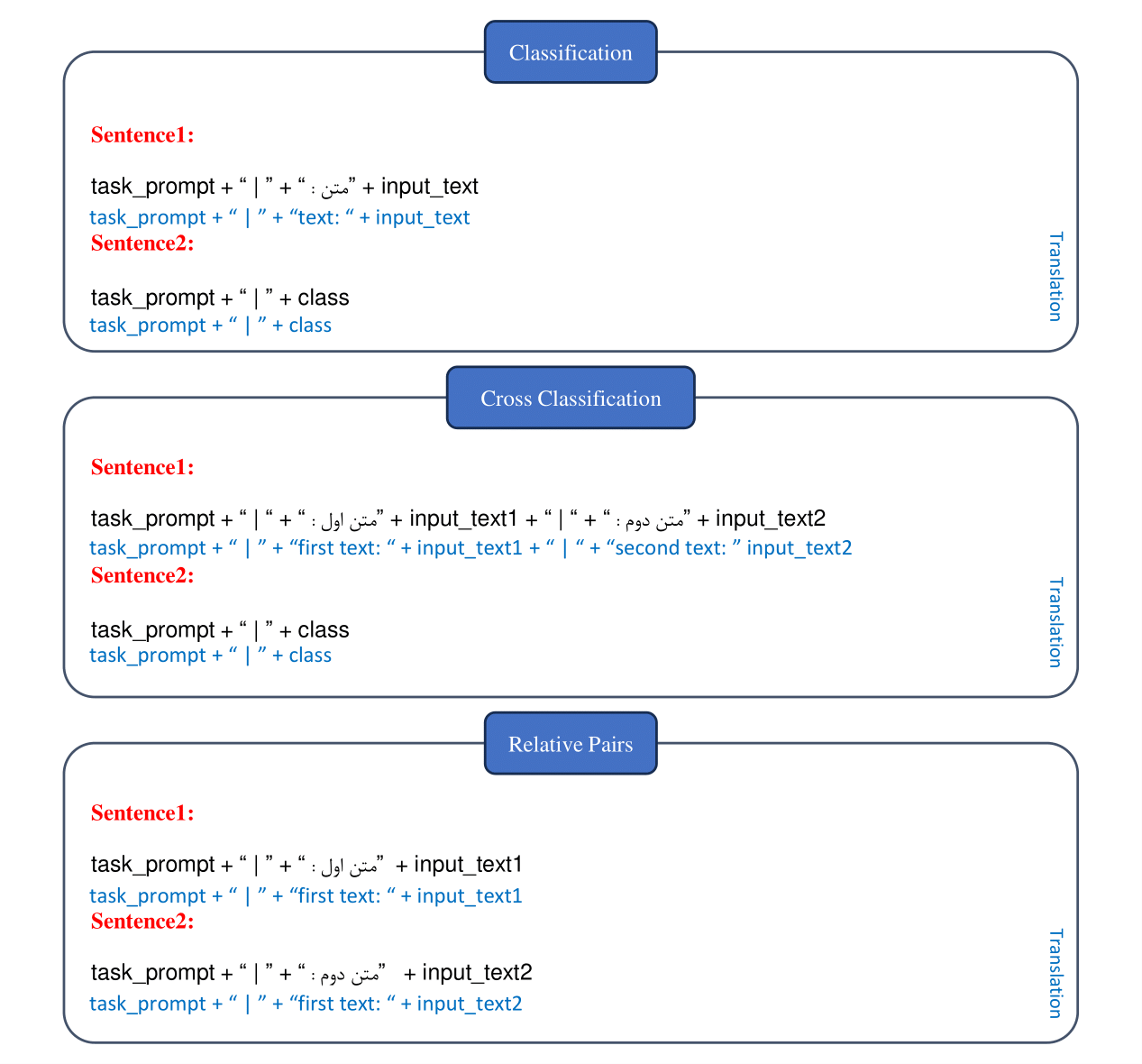}
\caption{The template of adding instructions to different data types used in Hakim. In general, data is categorized into three types: classification, cross-classification, and Relative Pairs types. Instructions and data pairs are added accordingly.}
\label{fig:inst_structure}
\end{figure}

\begin{figure*}[ht]
\centering
\includegraphics[width=\textwidth]{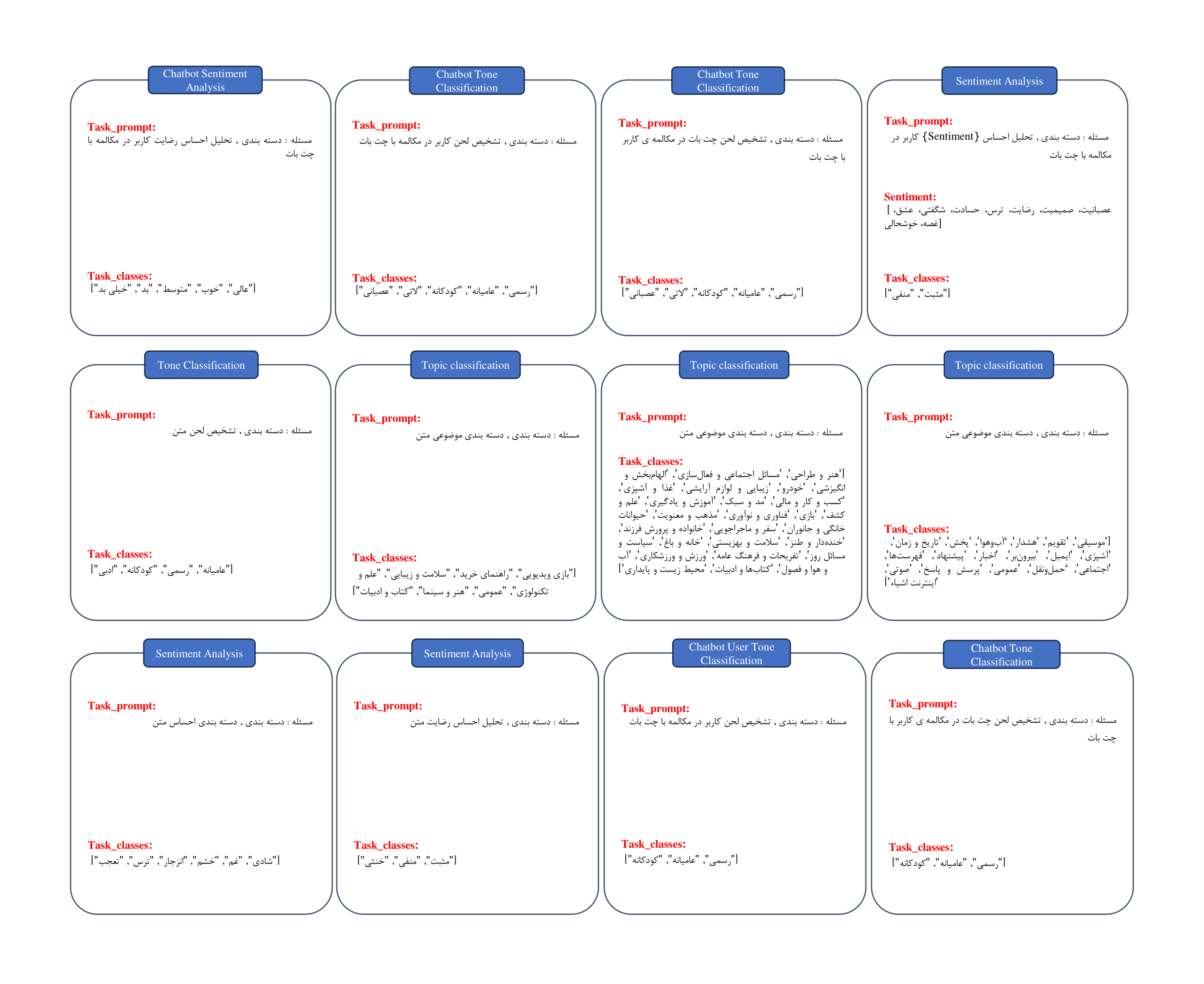}
\caption{The instructions employed for addressing classification tasks.}
\label{fig:Classification}
\end{figure*}

\begin{figure*}[ht]
\centering
\includegraphics[width=\textwidth]{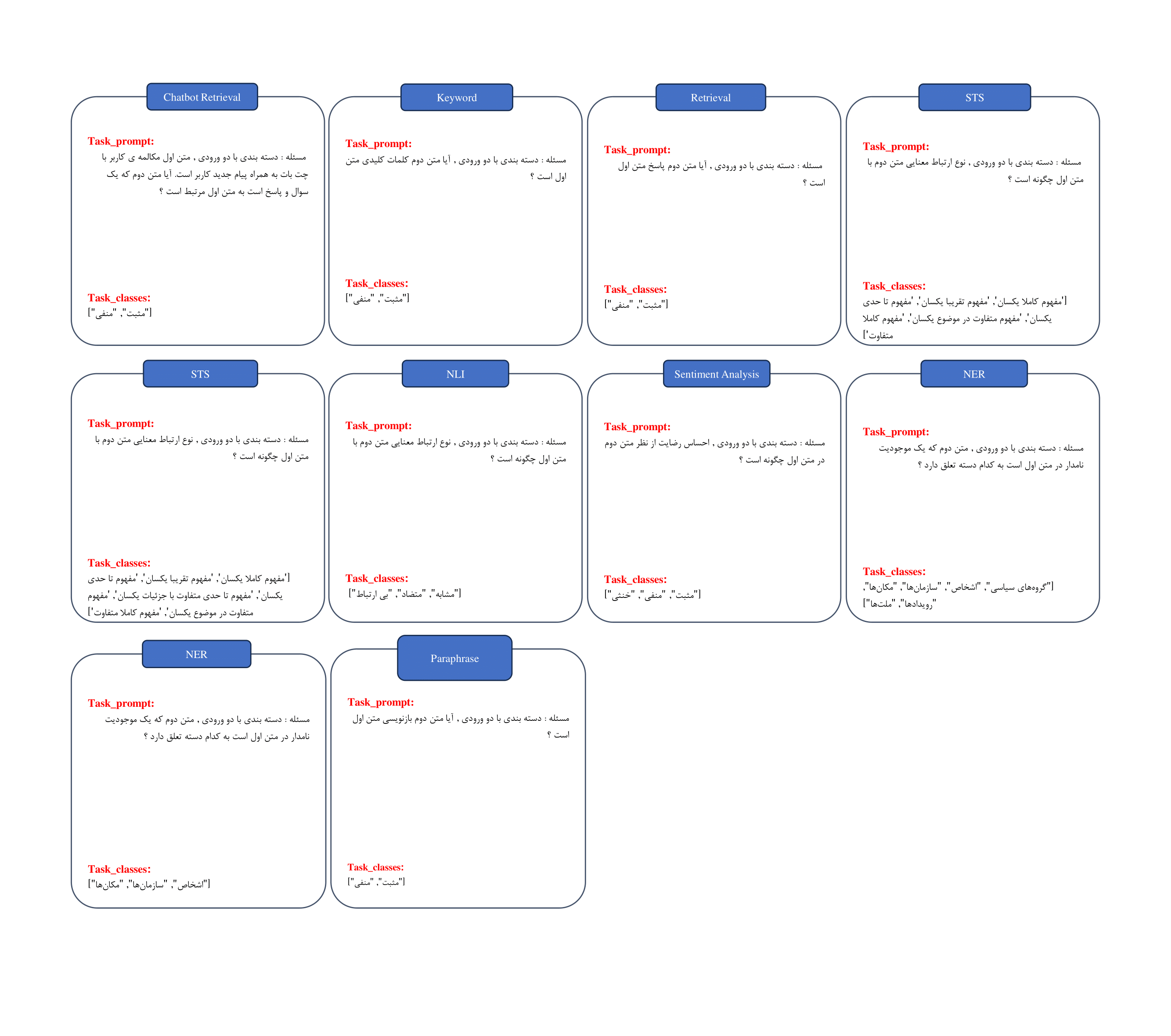}
\caption{The instructions employed for addressing cross classification tasks.}
\label{fig:Cross_Classification}
\end{figure*}

\begin{figure*}[ht]
\centering
\includegraphics[width=\textwidth]{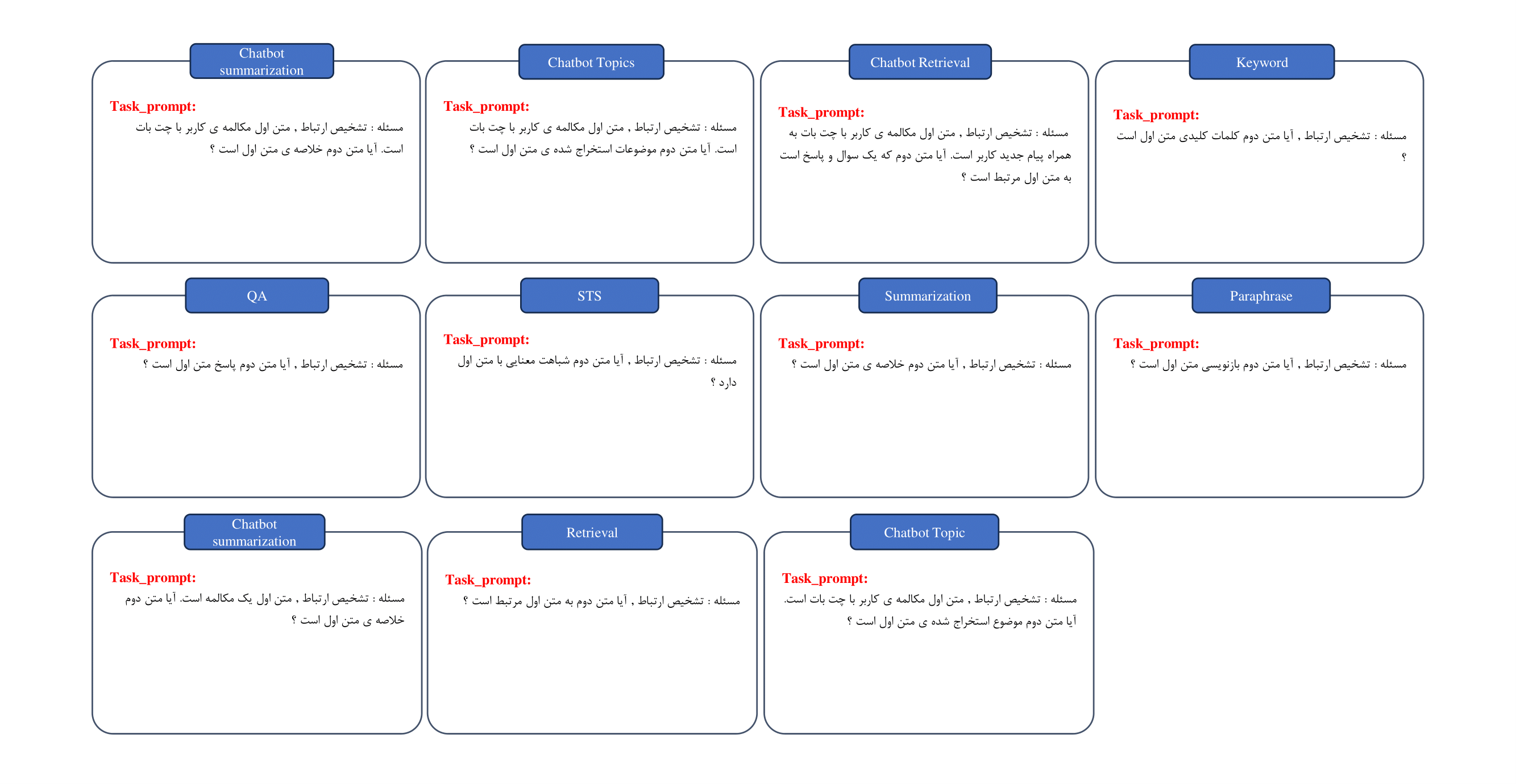}
\caption{The instructions employed for addressing Relative Pair tasks like retrieval, sts, and summarization.}
\label{fig:Other}
\end{figure*}

To optimize the model’s performance, we experimented with various instruction formulations during training. Each variant reflects a different strategy for instruction construction, and we refer to the resulting models as Hakim-inst1 through Hakim-inst5. Among these, Hakim-inst1 is the variant we designate as our main model, Hakim, throughout the paper.
\begin{itemize}
    \item Inst1 uses the full instruction templates introduced in Figure~\ref{fig:inst_structure} without modifications, and also the exact instructions used for each task is in Figure \ref{fig:Classification}, \ref{fig:Cross_Classification},and \ref{fig:Other}. This configuration serves as the baseline for our instruction-tuned model.
    \item Inst2 follows the same instruction format as shown in Figure~\ref{fig:inst_structure}, with two key exception: in classification and cross classification tasks, the list of class names is included right after task\_prompt in the query instruction.
    \item Inst3 provides a more concise version of the instructions used in Inst1, aiming to reduce verbosity while preserving task-relevant information.
    \item Inst4 uses the same abbreviated instructions as Inst3 but differs in its training regime. Unlike Inst2–Inst3, Inst4 is not exposed to any examples without instructions. Exposing the model to both instructed and non-instructed inputs during training enhances generalization. Thus, Inst4 intentionally omits this exposure to study its effect.
    \item Inst5 also uses the same abbreviated instructions as Inst1; however, no instruction is added to the second pair, except in Classification tasks.
    \item We also experiment with a No-Inst variant, which, unlike other models, is trained without any instruction and is only trained in the Relative Pair setting. This is because training in the Cross and Classification settings without an instruction is not meaningful.
\end{itemize}

As shown in Table~\ref{tab:instruction_results}, the model trained with Inst1 achieves higher accuracy compared to the other models. These results also indicate that incorporating instructions into the second pair, as opposed to the more common approach in Inst5—where the instruction is added only to the query—leads to improved accuracy.

\begin{table*}[h!]
\centering
\begin{adjustbox}{width=\textwidth}
\begin{tabular}{c|ccccccc|c}
\toprule
\textbf{Instruction Version} & Class. & Cluster. & PairClass. & Rerank. & Retriv. & STS & SumRet. & Ave.\\
\hline
        
        Inst2 & \textbf{84.66} & 66.94 & 84.91 & 70.52 & 40.04 & 74.11 & \textbf{86.15} & 72.48 \\
        Inst3 & 83.56 & 67.59 & 89.30 & 69.74 & 39.68 & 77.28 & 85.74 & 73.27 \\
        Inst4 & 84.25 & 66.87 & 87.87 & 69.71 & 37.74 & 77.09 & 85.67 & 72.74 \\
        Inst4 w.o. Inst & 74.67 & 57.70 & 87.87 & 69.80 & 38.27 & 78.22 & 78.52 & 69.29 \\
        Inst5 & 84.60 & 70.42 & 89.84 & 69.32 & 38.26 & 78.47 & 84.82 & 73.67 \\
        No-Inst & 63.03 & 54.51 & 88.80 & \textbf{70.78} & 37.72 & \textbf{78.79} & 83.41 & 68.10 \\
        Inst1 (Hakim) & 84.56 & \textbf{70.46} & 89.75 & 69.46 & \textbf{40.43} & 76.62 & 85.41 & \textbf{73.81} \\
        Inst1 w.o. Inst & 72.84 & 56.97 & \textbf{89.76} & 69.00 & 35.92 & 78.62 & 82.92 & 69.43 \\

\bottomrule
\end{tabular}
\end{adjustbox}
\caption{Performance comparison of different instruction-tuning variants across multiple tasks. Inst1 (Hakim) achieves the highest average score, demonstrating the effectiveness of full instruction templates.}
\label{tab:instruction_results}
\end{table*}

\subsubsection{RAG}
To evaluate the model’s capability in tasks related to Retrieval-Augmented Generation (RAG), we measured the performance of the Hakim model on a subset of the FaMTEB dataset consisting of RAG data. As presented in Table~\ref{tab:RAG_accuracy}, fine-tuning the model on this dataset led to a significant improvement in accuracy, indicating its effectiveness in handling dialogue-based retrieval tasks.

\begin{figure}[ht]
\centering
\begin{subfigure}[b]{0.48\columnwidth}
    \centering
    \includegraphics[width=\linewidth]{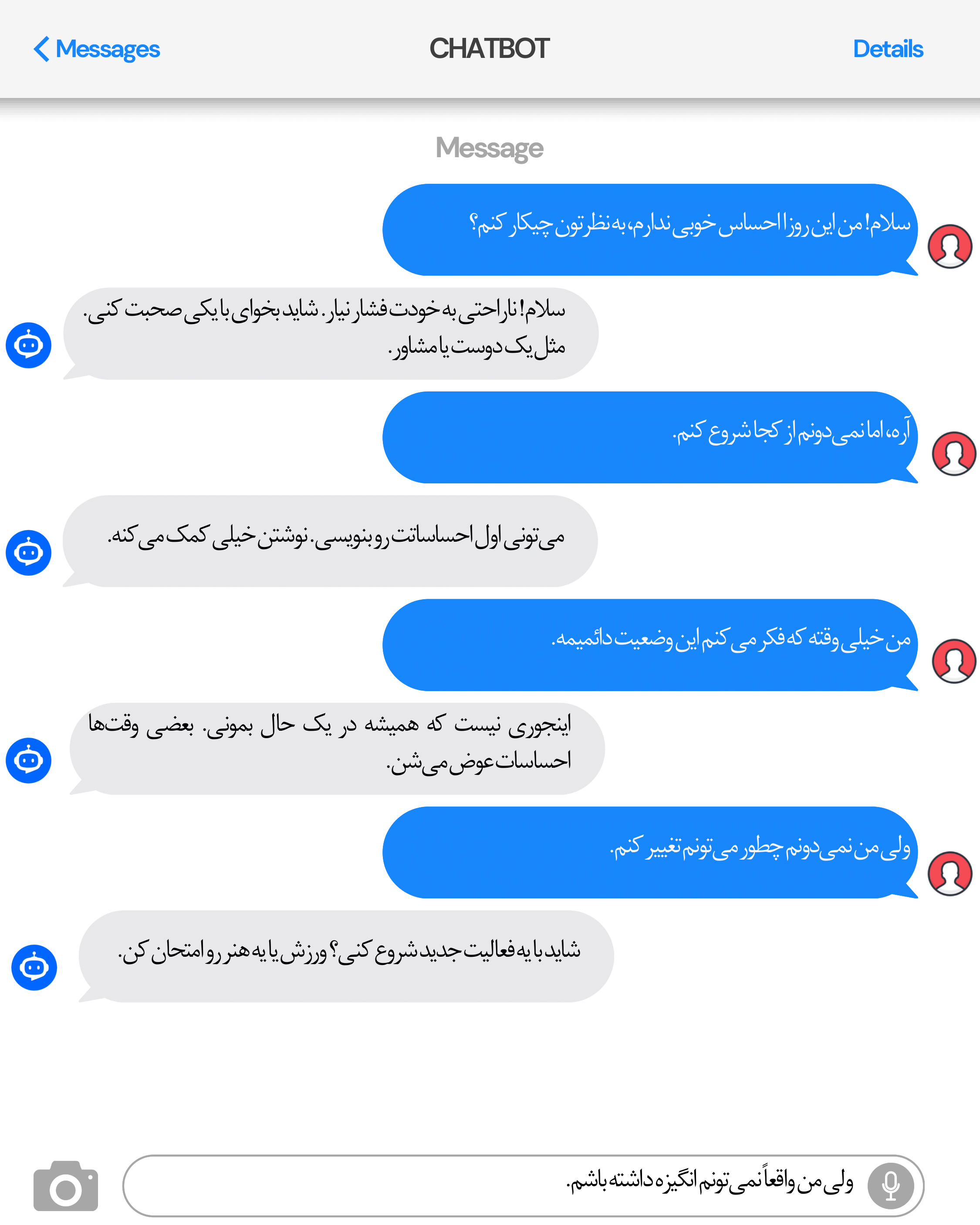}
    \caption{Chat}
    \label{fig:chat}
\end{subfigure}
\hfill
\begin{subfigure}[b]{0.48\columnwidth}
    \centering
    \includegraphics[width=\linewidth]{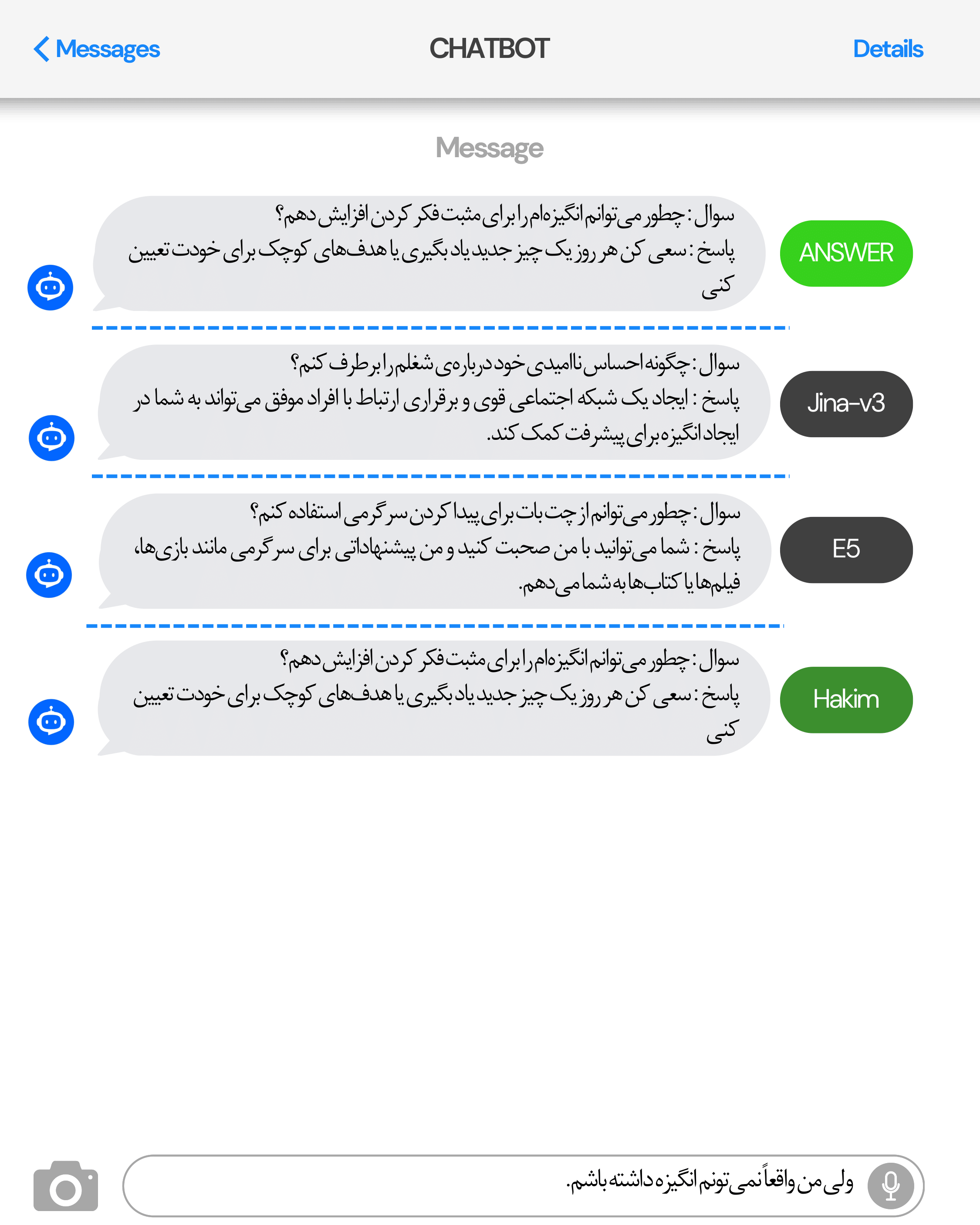}
    \caption{Answers}
    \label{fig:answer}
\end{subfigure}
\caption{Figure (a) illustrates a chat between a user and a chatbot, where the user asks a follow-up question that requires understanding of the prior conversation. Figure (b) presents the correct answer along with the responses retrieved by different models through a question-answering index. The Jina and E5 models return incorrect question-answer pairs, whereas the Hakim model successfully retrieves the correct answer.}
\label{fig:chat_answer}
\end{figure}

Additionally, Figure~\ref{fig:chat_answer} illustrates a qualitative example from the SynPerChatbotRAGFAQRetrieval dataset from FaMTEB benchmark. In this case, the user query requires understanding the preceding chat history to retrieve a relevant response. We compare the outputs of the Hakim with Jina and multilingual-e5. The results show that the Hakim model successfully retrieves the appropriate answer, demonstrating its ability to capture context and perform effective conversational retrieval.

\begin{table*}[h!]
\centering
\begin{adjustbox}{width=\textwidth}
\begin{tabular}{l|c|c|c|c}
\hline
\textbf{Dataset} & \textbf{Hakim} & \textbf{Hakim No-Inst} & \textbf{BGE-m3} & \textbf{Jina-embeddings-v3} \\
\hline
SynPerChatbotRAGFAQRetrieval & 54.70 & \textbf{58.19} & 32.03 & 47.45 \\
SynPerChatbotRAGFAQPC & \textbf{93.39} & 89.63 & 64.42 & 62.05 \\
SynPerChatbotConvSAClassification & \textbf{89.84} & 78.84 & 61.02 &  71.57 \\
SynPerChatbotTopicsRetrieval & \textbf{52.15} & 18.80 & 19.18 & 18.75 \\
SynPerChatbotRAGTopicsRetrieval & \textbf{50.02} & 28.37 & 19.91 & 24.26 \\

\hline
\end{tabular}
\end{adjustbox}
\caption{Evaluation of retrieval-augmented generation performance on FaMTEB subsets. Hakim outperforms baseline models, demonstrating superior contextual understanding in dialogue-based retrieval tasks.}
\label{tab:RAG_accuracy}
\end{table*}

\subsection{Cross Classification}
For training on the cross classification task, we set aside a portion of the dataset for evaluation. Table~\ref{tab:cross_results} reports the accuracy of the Hakim model compared to other baseline models on this dataset. As shown, the model is capable of handling tasks in a cross formulation, allowing it to address problems that were previously not directly solvable using traditional text embedding models. For instance, semantic textual similarity (STS) can now be performed using the Hakim model by encoding a pair of sentences within an instruction, and associating the pair with varying similarity scores, effectively transforming the task into a classification or regression problem within the cross classification framework.

\begin{table*}[ht]
\centering
\begin{adjustbox}{width=\textwidth}
\begin{tabular}{c|cccccccccccc}
\textbf{Model} & 
\rotatebox{90}{CExappc} & 
\rotatebox{90}{ChatbotRagFAQ} & 
\rotatebox{90}{FarsiParaphraseDetection} & 
\rotatebox{90}{KeywordAndToneKeyword} & 
\rotatebox{90}{Farstail} & 
\rotatebox{90}{ParsABSA} & 
\rotatebox{90}{ParsinluEntail} & 
\rotatebox{90}{ParsinluQueryParaph} & 
\rotatebox{90}{ParsiTwiNER} & 
\rotatebox{90}{STSSyn} & 
\rotatebox{90}{Wikiann} & 
\rotatebox{90}{SyntheticQAFa} \\
\hline

Hakim & \textbf{99.45} & \textbf{92.58} & \textbf{97.22} & \textbf{98.84} & \textbf{93.79} & \textbf{89.26} & \textbf{57.12} & \textbf{83.58} & \textbf{94.23} & \textbf{80.6} & \textbf{95.74} & \textbf{96.87} \\
Alibaba-NLP-gte-multilingual-base & 87.63 & 51.91 & 59.16 & 50.27 & 33.6 & 49.34 & 34.09 & 52.8 & 30.52 & 30.54 & 66.18 & 50.62 \\
BAAI-bge-m3 & 93.34 & 51.79 & 51.79 & 51.57 & 34.07 & 54.8 & 35.69 & 53.84 & 31.24 & 31.42 & 63.73 & 52.08 \\
intfloat-multilingual-e5-large & 94.5 & 52.47 & 67.87 & 51.11 & 33.75 & 48.38 & 36.08 & 53.0 & 29.91 & 33.31 & 69.75 & 51.4 \\
tooka-sbert & 86.18 & 53.4 & 53.81 & 50.34 & 33.64 & 54.84 & 35.85 & 51.09 & 23.06 & 30.36 & 55.57 & 50.5 \\
sentence-transformers & 85.69 & 52.47 & 56.0 & 50.43 & 33.66 & 45.94 & 34.49 & 53.23 & 31.55 & 29.58 & 60.98 & 50.48 \\
\bottomrule
\end{tabular}
\end{adjustbox}
\caption{Accuracy comparison of Hakim and baseline models on 12 Persian cross classification tasks, highlighting the impact of instruction tuning.}
\label{tab:cross_results}
\end{table*}

\subsection{HakimBERT Analysis}
To evaluate our base model, we conducted a series of experiments. Our first experiment compares the training data used for HakimBERT with the Targoman dataset. In this experiment, we train two BERT models with identical tokenizers and hyperparameters on our dataset and the Targoman dataset, respectively. As shown in Table~\ref{tab:targoman_ours}, the model trained on our dataset consistently outperforms the one trained on Targoman across most tasks.

We also assess the effectiveness of our tokenizer by training two models on the same dataset: one using our tokenizer and the other using the TookaBERT tokenizer. The results, presented in Table~\ref{tab:tokenizers}, demonstrate that, except for two tasks—where the models perform comparably, the model trained with the Hakim tokenizer achieves higher accuracy in all other tasks.

To further investigate the effectiveness of RetroMAE training, we perform a second-stage pretraining (unsupervised finetuning) on two models: one using the original HakimBERT and the other using HakimBERT pretrained with the RetroMAE objective. As shown in Table~\ref{tab:retro_results}, we observe accuracy improvements across most datasets when using the RetroMAE-enhanced model, indicating the benefit of incorporating the RetroMAE objective into the training process.

\begin{table*}[ht]
\centering
\label{tab:comparison2}
\resizebox{\textwidth}{!}{%
\begin{tabular}{l|cc|cc|cc|cc}
\toprule
& \multicolumn{2}{c|}{\textbf{SA}} & \multicolumn{2}{c|}{\textbf{QA}} & \multicolumn{2}{c|}{\textbf{NER}} & \multicolumn{2}{c}{\textbf{NLI}}  \\
\cmidrule(r){2-3} \cmidrule(r){4-5} \cmidrule(r){6-7} \cmidrule(r){8-9}
\textbf{Model} &  \textbf{DeepSentiPers} & \textbf{MirasOpinion} & \textbf{PQuAD} & \textbf{PCoQA} & \textbf{ParsTwiner} & \textbf{MULTICONER V2} & \textbf{FarsTail} & \textbf{ParsiNLU QP} \\
\midrule
Targoman & 81.96 & \textbf{85.10} & 87.87 & 45.29 & \textbf{85.42} & 59.63 & 81.14 & 80.34 \\
Ours  & \textbf{82.17} & 84.23 & \textbf{88.47} & \textbf{47.83} & 83.69 & \textbf{60.66} & \textbf{85.40} & \textbf{80.39} \\ 

\bottomrule
\end{tabular}%
}
\caption{Comparison of BERT-based models trained on our dataset vs. the Targoman dataset across eight Persian NLU tasks, showing the impact of training data on performance.}
\label{tab:targoman_ours}
\end{table*}

\begin{table*}[ht]
\centering
\label{tab:comparison3}
\resizebox{\textwidth}{!}{%
\begin{tabular}{l|cc|cc|cc|cc}
\toprule
& \multicolumn{2}{c|}{\textbf{SA}} & \multicolumn{2}{c|}{\textbf{QA}} & \multicolumn{2}{c|}{\textbf{NER}} & \multicolumn{2}{c}{\textbf{NLI}}  \\
\cmidrule(r){2-3} \cmidrule(r){4-5} \cmidrule(r){6-7} \cmidrule(r){8-9}
\textbf{Model} &  \textbf{DeepSentiPers} & \textbf{MirasOpinion} & \textbf{PQuAD} & \textbf{PCoQA} & \textbf{ParsTwiner} & \textbf{MULTICONER V2} & \textbf{FarsTail} & \textbf{ParsiNLU QP} \\
\midrule
TookaBERT  & 81.80 & 84.74 & \textbf{88.46} & \textbf{45.26} & 84.65 & 60.68 & 85.23 & 81.15 \\
Ours & \textbf{82.99} & \textbf{85.04} & 88.31 & 45.03 & \textbf{86.11} & \textbf{61.42} & \textbf{86.82} & \textbf{83.53} \\ 

\bottomrule
\end{tabular}%
}
\caption{Performance comparison of models trained with different tokenizers (Tooka vs. Ours) on eight Persian NLU tasks, showing the effect of tokenizer choice on downstream performance.}
\label{tab:tokenizers}
\end{table*}

\begin{table*}[ht]
\centering
\begin{adjustbox}{width=\textwidth}
\begin{tabular}{c|ccccccc|c}

\textbf{Prompt Version} & Class. & Cluster. & PairClass. & Rerank. & Retriv. & STS & SumRet. & Ave.\\
\hline

    Hakim-unsup w.o. Retro & 58.75 & 60.12 & 85.75 & 65.81 & 35.56 & 78.90 & 50.85 & 62.25 \\
    Hakim-unsup & \textbf{60.65} & \textbf{58.89} & \textbf{86.41} & \textbf{67.56} & \textbf{37.71} & \textbf{79.36} & \textbf{61.34} & \textbf{64.56} \\

\end{tabular}
\end{adjustbox}
\caption{Comparison of Hakim-unsup model performance with and without RetroMAE enhancement across seven faMTEB tasks. Results show consistent gains from incorporating RetroMAE.}
\label{tab:retro_results}
\end{table*}

\section{Conclusion}
In this work, we introduced Hakim, a state-of-the-art Persian text embedding model designed to address the longstanding underrepresentation of Persian in large-scale NLP research. By constructing and utilizing three new high-quality datasets—Corpesia, Pairsia-unsup, and Pairsia-sup—we established a comprehensive training pipeline that supports both unsupervised and supervised learning paradigms. Our model not only surpasses previous approaches on the FaMTEB benchmark with an 8.5\% performance improvement, but also demonstrates robust capabilities in downstream applications such as chatbots, retrieval-augmented generation (RAG), and a novel cross-classification task. The integration of instruction-based supervision and domain-specific corpus construction has proven effective in capturing the linguistic nuances of Persian, paving the way for more accurate and contextually aware embeddings. Overall, our contributions lay a strong foundation for further advances in Persian NLP, and we anticipate that our works in this paper will support a wide range of future research and practical applications.

\FloatBarrier
\bibliography{acl_latex}

\end{document}